\documentclass{article}

    \PassOptionsToPackage{numbers, compress}{natbib}

\usepackage[final]{neurips_2021}

\usepackage[utf8]{inputenc} 
\usepackage[T1]{fontenc}    
\usepackage{hyperref}       
\usepackage{url}            
\usepackage{booktabs}       
\usepackage{amsfonts}       
\usepackage{nicefrac}       
\usepackage{microtype}      
\usepackage{xcolor}         
\usepackage{graphicx}
\usepackage{footmisc}
\usepackage{booktabs}
\usepackage{tablefootnote}
\usepackage{amsmath}
\usepackage{caption}
\usepackage{graphicx}
\usepackage{placeins}
\usepackage{pgfplots}
\usepackage{pgfplotstable}
\usepackage{tikz}
\usepackage{adjustbox}
\usepackage{amsmath}
\usepackage{float}
\usepackage{footnote}
\usepackage{etoolbox}
\usepackage{array, booktabs, makecell}
\usepackage{subcaption}
\usepackage{nameref}

\title{Dynamic-TinyBERT: Boost TinyBERT's Inference Efficiency by Dynamic Sequence Length}

\author{%
 Shira Guskin \\
  Intel Labs \\
  \texttt{shira.guskin@intel.com}
  \And
  Moshe Wasserblat \\
  Intel Labs \\
  \texttt{moshe.wasserblat@intel.com}
  \And
  Ke Ding \\
  Intel \\
  \texttt{ke.ding@intel.com}
  \And
  Gyuwan Kim \\
  University of California, Santa Barbara \\
  \texttt{gyuwankim@ucsb.edu}
}

\begin{document}

\maketitle

\begin{abstract}
Limited computational budgets often prevent transformers from being used in production and from having their high accuracy utilized. TinyBERT~\cite{Jiao2020TinyBERTDB} addresses the computational efficiency by self-distilling BERT~\cite{Devlin2019BERTPO} into a smaller transformer representation having fewer layers and smaller internal embedding. However, TinyBERT's performance drops when we reduce the number of layers by 50\%, and drops even more abruptly when we reduce the number of layers by 75\% for advanced NLP tasks such as span question answering.  Additionally, a separate model must be trained for each inference scenario with its distinct computational budget. 
In this work we present \textit{Dynamic-TinyBERT}, a TinyBERT model that utilizes sequence-length reduction and Hyperparameter Optimization for enhanced inference efficiency per any computational budget. Dynamic-TinyBERT is trained only once, performing on-par with BERT and achieving an accuracy-speedup trade-off superior to any other efficient approaches (up to 3.3x with <1\% loss-drop). Upon publication, the code to reproduce our work will be open-sourced.

\end{abstract}

\section{Introduction}
\label{sec:intro}
In recent years, increasingly large Transformer-based models such as BERT~\cite{Devlin2019BERTPO}, RoBERTa~\cite{Liu2019RoBERTaAR} and GPT-3~\cite{Brown2020LanguageMA} have demonstrated remarkable state-of-the-art (SoTA) performance in many Natural Language Processing (NLP) and Computer Vision (CV) tasks and have become the de-facto standard. 
However, those models are extremely inefficient; they require massive computational resources and large amounts of data as basic requirements for training and deploying. This severely hinders the scalability and deployment of AI-based systems across the industry.

One highly effective method for improving efficiency is Knowledge Distillation~\citep{Ba2014DoDN,Hinton2015DistillingTK}, in which the knowledge of a large model defined as the teacher is transferred into a smaller more efficient model defined as the student. TinyBERT~\cite{Jiao2020TinyBERTDB} stands out with its superior accuracy-speed-size tradeoff, introducing transformer distillation — a novel distillation method specially designed for transformer-based models — that transfers the knowledge residing in the hidden states and attention matrices of BERT using a two-stage learning framework that captures both the general domain and the task-specific knowledge of BERT.

Knowledge distillation has shown promising results for reducing the number of parameters, with, however, several caveats: First, a drop in accuracy (>1\%) and a still limited speed-up/latency gain, specifically in challenging NLP tasks such as QA; for example, DistilBERT~\cite{Sanh2019DistilBERTAD} produces a 1.7x speed-up albeit with a 3\% accuracy drop on SQuAD 1.1. Secondly, in many cases the target computational budget (HW type, memory size, latency constraints, etc.) is not given at the time of training. This implies that a separate student model must be trained for each applicable inference scenario and its distinct computational budget.

Recent studies have attempted to address these concerns by proposing dynamic transformers. Among these, Funnel Transformer~\cite{Dai2020FunnelTransformerFO} successfully reduced the sequence length, however its size is fixed and not designed to control efficiency. DynaBERT~\cite{Hou2020DynaBERTDB} can run at adaptive width (the number of attention heads and intermediate hidden dimensions) and depth but requires training a separate model for each computational budget. Schwartz et al.~\cite{Schwartz2020TheRT} proposed a transformer-based architecture that uses an early exit strategy based on the confidence of the prediction in each layer. The method worked well when combined with TinyBERT on classification tasks (e.g. IMDB) but applying it to “difficult” NLP tasks is still being actively researched.

PoWER-BERT~\cite{Goyal2020PoWERBERTAB} progressively reduces sequence length by eliminating word-vectors based on the attention values as they pass the layers. However, it needs to retrain for a given computational budget and is not applicable to a wider range of NLP tasks such as span-based question answering. 

LAT~\cite{kim-cho-2021-length} extends PoWER-BERT by introducing LengthDrop, a structured variant of dropout for training a single model that can be adapted during the inference stage to meet any given efficiency target with its maximized accuracy-efficiency tradeoff achieved by a multi-objective evolutionary search. LAT also includes a Drop-and-Restore process that extends the applicability of PoWER-BERT beyond classification, to a wider range of NLP tasks such as span-based question answering.  LAT was shown to work quite well when applied to BERT and to DistilBERT on a diverse set of NLP tasks, including SQuAD1.1~\cite{Rajpurkar2016SQuAD1Q} and shows a superior accuracy-efficiency tradeoff in the inference time, compared to other existing approaches.

In this paper we present Dynamic-TinyBERT — a TinyBERT model that supports sequence length reduction along the layers for faster inference. Dynamic-TinyBERT performs on-par with BERT with 60\% of parameters and demonstrates an accuracy-efficiency tradeoff that is superior to any other efficiency approach (up to x3.3 speedup with <1\% loss) on the challenging SQuAD1.1 benchmark. Following the concept presented by LAT~\cite{kim-cho-2021-length}, it provides a wide range of accuracy-efficiency tradeoff points while alleviating the need to retrain it for each point along the accuracy-efficiency curve. Furthermore, we explore the implications of incorporating LAT's~\cite{kim-cho-2021-length} LengthDrop method into the training of Dynamic-TinyBERT for increasing the robustness of Dynamic-TinyBERT to sequence-length reduction.

\begin{figure}[t]
    \centering
    \includegraphics[width=14cm, height=5.5cm]{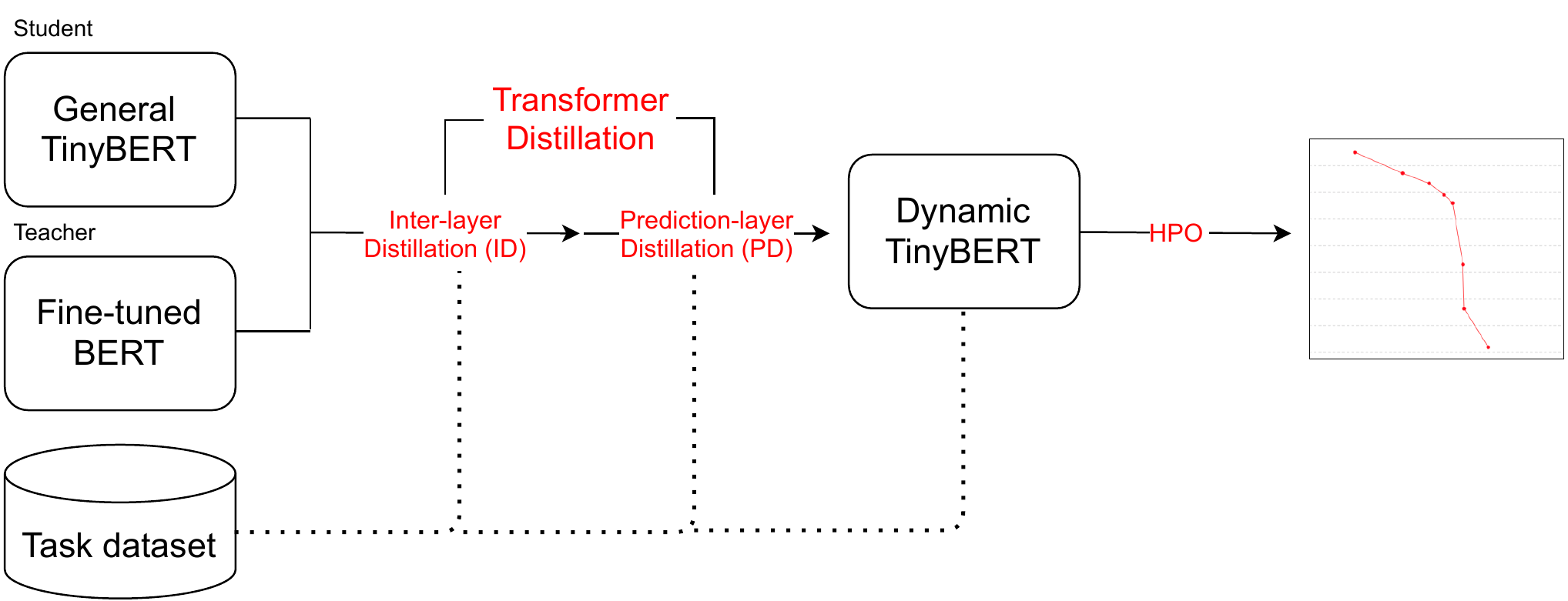}
    \caption{Dynamic-TinyBERT training process}
    \label{fig:training_fig}
\end{figure}

\section{Method}

Dynamic-TinyBERT is trained slightly different than TinyBERT (see section~\ref{sec:training}), achieving better accuracy, and is run with the Drop-and-Restore method proposed by LAT~\cite{kim-cho-2021-length}: word-vectors are eliminated in the encoder layers according to a given length configuration (a list of sequence-lengths per layer), then brought back in the last hidden layer. We use Drop-and-Restore only during the inference stage. To adapt to any computational budget with best performance we use Hyperparameter Optimization as described with details in section~\ref{sec:hpo}. 

\subsection{Training}
\label{sec:training}
The training procedure of Dynamic-TinyBERT is illustrated in Figure~\ref{fig:training_fig}. We start with a pre-trained general-TinyBERT student, which was trained to learn the general knowledge of BERT using the general-distillation method presented by TinyBERT. We perform transformer distillation from a fine-tuned BERT teacher to the student, following the same training steps used in the original TinyBERT: (1) \textbf{intermediate-layer distillation (ID)} — learning the knowledge residing in the hidden states and attentions matrices, and (2)  \textbf{prediction-layer distillation (PD)} — fitting the predictions of the teacher as in Hinton et al.~\citep{Hinton2015DistillingTK}. Unlike the original TinyBERT, we use the original task dataset without performing data augmentation and we perform the PD step for a larger number of epochs.

\subsection{Hyperparameter Optimization}
\label{sec:hpo}
We borrow LAT's proposal to use search over length-configurations for optimizing the performance of Dynamic-TinyBERT for any possible target computational budget. Differently from the evolutionary-search method used by LAT for this purpose, we use Hyperparameter Optimization (HPO) over length-configurations. In our experiment we use SigOpt which is a leading Bayesian Optimization software service provider\footnote{\url{app.sigopt.com}}. The HPO search space contains 6 variables x0 to x5, representing the sequence length for each encoder layer. 5 linear constraints are applied to ensure the sequence length from previous layer is always greater or equal to the current one.
Thanks to SigOpt’s multi-metrics optimization feature, we are able to design two optimization targets during the HPO process. One is F1 score to maximize, another is evaluation time to minimize. The multi-metrics optimization result is rendered as a Pareto-front for user to select best parameter set according to the actual use case.
We set the optimization budget as 150 experiments and run parallel experiment in order to improve the efficiency.
 
Comparing to the naive evolutionary-search used in LAT, the SigOpt HPO gives us much better search efficiency with a much smaller budget yet still gives us the optimal Pareto-front. This is mainly because our 6 variable search space falls into the region where Bayesian optimization can perform really well.

\subsection{Training with LengthDrop}
LAT~\cite{kim-cho-2021-length} introduced a method called LengthDrop (LD) for training a model to be robust to different length configurations that are given at inference time. Training with LengthDrop consists of reducing the sequence length by a random proportion at each
layer during the training phase. LAT showed good results on BERT~\cite{Devlin2019BERTPO} and DistilBERT~\cite{Sanh2019DistilBERTAD} for training a model with LengthDrop using inplace distillation and a sandwich rule, as follows: randomly-sampled sub-models with length reduction (sandwiches) learn to mimic the predictions of the full model while simultaneously the full model is being fine-tuned for the downstream task.  

We explore whether LengthDrop (LD) improves upon the Dynamic-TinyBERT's robustness to length reduction by testing several variants of training procedures incorporating LengthDrop, which differ by the steps in which LengthDrop is added and by the number of epochs of the different stages. Below we describe the tested procedures. For convenience we use the following pipeline notation for each procedure: (1) M\textsubscript{1},E\textsubscript{1},L\textsubscript{1} $\rightarrow$ (2) M\textsubscript{2},E\textsubscript{2},L\textsubscript{2} $\rightarrow$ (3) etc., where M\textsubscript{i} is one of \{ID, PD\} methods, E\textsubscript{i} represents the number of epochs and L\textsubscript{i} is either True (T) or False (F) depending on whether LengthDrop training was added in that step. Each step's output is used as the student model of the next step, and the teacher used in all of the procedures is a fine-tuned BERT-base except in Dynamic-TinyBERT\textsubscript{w/ LD}v4 where the teacher is a fine-tuned BERT which was trained with LengthDrop.\newline

\textbf{Dynamic-TinyBERT\textsubscript{w/ LD}naive:} (1) ID,20,F $\rightarrow$ (2) PD,10,F $\rightarrow$  PD,10,T. This procedure is the naive implementation of the procedure used in LAT, where the model is first being trained without LengthDrop, then trained with LengthDrop for additional number of epochs. \newline

\textbf{Dynamic-TinyBERT\textsubscript{w/ LD}v1:} (1) ID,20,F $\rightarrow$ (2) PD,20,T.\newline
\textbf{Dynamic-TinyBERT\textsubscript{w/ LD}v2:} (1) ID,10,F $\rightarrow$ (2) PD,3,F $\rightarrow$ (3) PD,10,T.\newline
\textbf{Dynamic-TinyBERT\textsubscript{w/ LD}v3:} (1) ID,20,T $\rightarrow$ (2) PD,10,T.\newline
\textbf{Dynamic-TinyBERT\textsubscript{w/ LD}v4:} Here the teacher is trained by: (1) BERT fine-tuning,2,F $\rightarrow$ (2) BERT fine-tuning,5,T and the model is trained by: (1) ID,20,F $\rightarrow$ (2) PD,10,F under the supervision of the new teacher.

\section{Experiments}

\begin{figure}[t]

\resizebox{0.9\textwidth}{!}{

\begin{tikzpicture}
\definecolor{s1}{RGB}{197, 90, 17}
\definecolor{clr_orange}{RGB}{255, 127, 0}
\definecolor{clr_green}{RGB}{31, 182, 83}
\definecolor{clr_purple}{RGB}{182, 83, 204}
\definecolor{clr_fuchsia}{RGB}{145, 92, 130}
\begin{axis}[
    width=\linewidth,
    xlabel={Speedup relative to BERT-base},
    ylabel={Accuracy[\%]},
    xmin=1, xmax=5.5,
    ymajorgrids=true,
    grid style=dashed,
    legend columns=1,
    legend pos=north east,
    legend style={at={(0.60,0.99)},anchor=north west},
    xtick pos=left, ytick pos=left,
    xtick={1, 1.5,2, 2.5, 3, 3.5, 4, 4.5, 5, 5.5},
    ytick={85,85.5,86,86.5,87,88, 88.5},
    extra y ticks = {87.5},
    extra y tick style = {brown},
    extra y tick labels = {{BERT 1\% 87.5}},
    yticklabels={85,85.5,86,86.5,87,88, {BERT 88.5}},
    xticklabels={1,1.5,2, 2.5, 3, 3.5, 4, 4.5, 5, 5.5},
    typeset ticklabels with strut,
    enlarge x limits=false,
    scale only axis]
    \addplot[red, mark=square*] table [x=speedup, y=f1, col sep=comma] {Dynamic-TinyBERT.csv};
    \addplot[pink] table [x=speedup, y=f1, col sep=comma] {Dynamic-TinyBERT_naive.csv};
    \addplot[black ,mark=*, mark size=3pt, nodes near coords={TinyBERT}] coordinates{(2, 87.5)}; 
    \addplot[black ,mark=*, mark size=3pt, nodes near coords={DistilBERT}] coordinates{(1.7, 85.8)}; 
    \addplot[black ,mark=*, mark size=3pt] coordinates{(1, 88.5)};
    \addplot[brown ,mark=*, mark size=3pt] coordinates{(1, 87.5)};
    \addplot[mark=none, black] coordinates {(1,88.5) (5.5,88.5)}; 
    \addplot[mark=none, brown] coordinates {(1,87.5) (5.5,87.5)}; 
    \legend{Dynamic-TinyBERT, Dynamic-TinyBERT\textsubscript{w/ LD}naive}
\end{axis}

\end{tikzpicture}
}

\hfill

\vspace{2mm}
\caption{Pareto-curves of Dynamic-TinyBERT which was not trained with LengthDrop vs. Dynamic-TinyBERT\textsubscript{w/ LD}naive which was trained with LengthDrop}\label{fig:dynamic}


\end{figure}
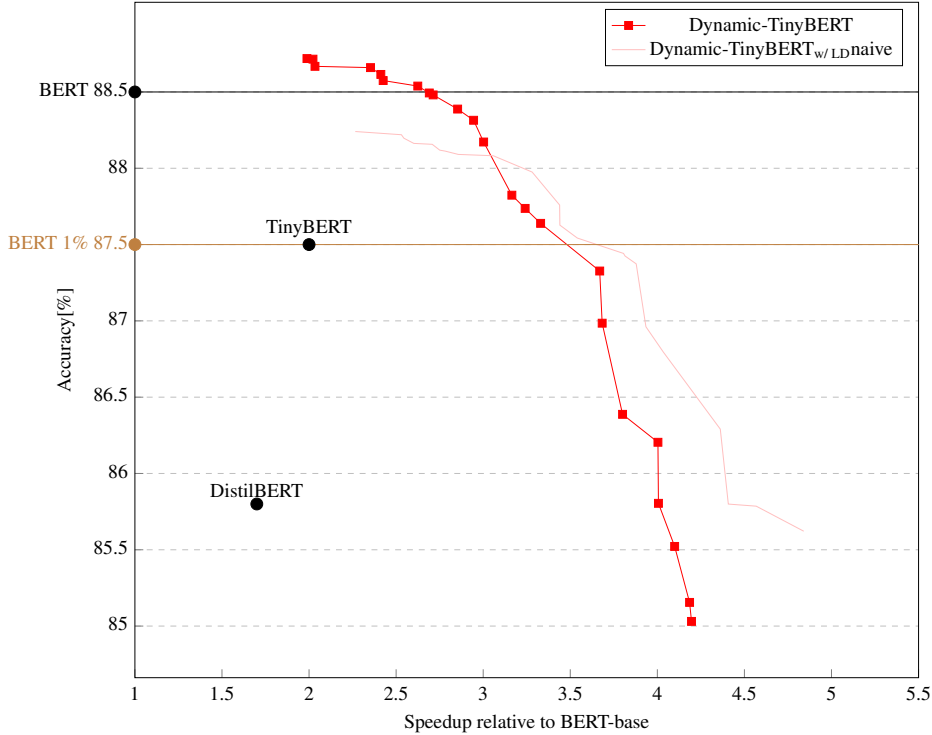

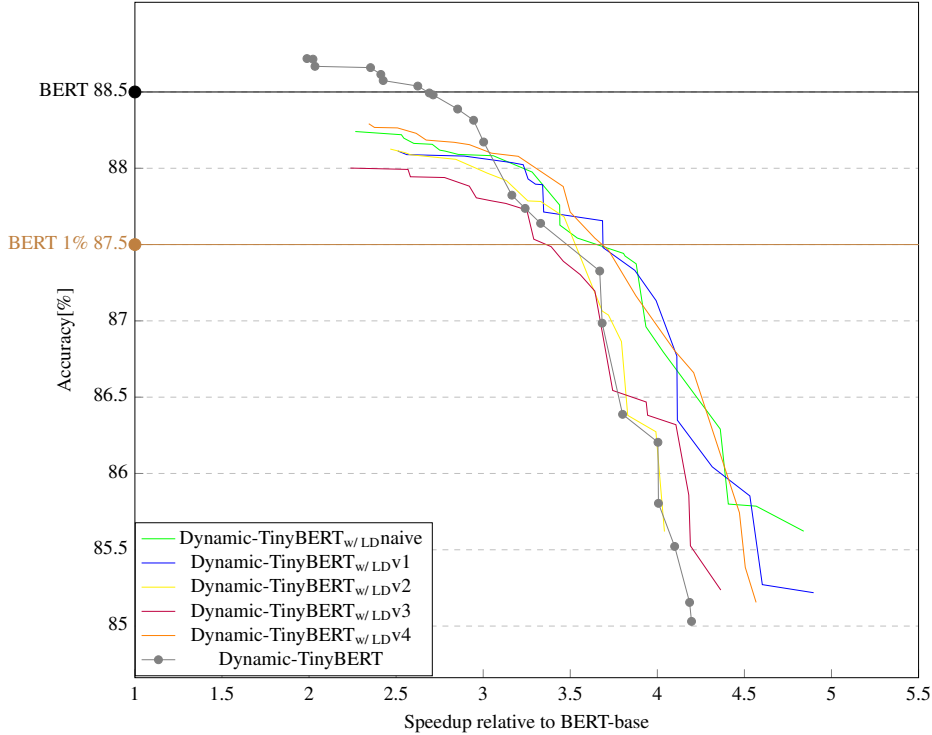
\begin{figure}[t]
\resizebox{0.9\textwidth}{!}{
\begin{tikzpicture}

\begin{axis}[
    width=\linewidth,
    xlabel={Speedup relative to BERT-base},
    ylabel={Accuracy[\%]},
    xmin=1, xmax=5.5,
    ymajorgrids=true,
    grid style=dashed,
    legend columns=1,
    legend style={at={(0,0.23)},anchor=north west},
    xtick pos=left, ytick pos=left,
    xtick={1, 1.5,2, 2.5, 3, 3.5, 4, 4.5, 5, 5.5},
    ytick={85,85.5,86,86.5,87,88, 88.5},
    extra y ticks = {87.5},
    extra y tick style = {brown},
    extra y tick labels = {{BERT 1\% 87.5}},
    yticklabels={85,85.5,86,86.5,87,88, {BERT 88.5}},
    xticklabels={1,1.5,2, 2.5, 3, 3.5, 4, 4.5, 5, 5.5},
    typeset ticklabels with strut,
    enlarge x limits=false, 
    scale only axis]
    \addplot[green] table [x=speedup, y=f1, col sep=comma] {Dynamic-TinyBERT_naive.csv};
    \addplot[blue] table [x=speedup, y=f1, col sep=comma] {Dynamic-TinyBERTv1.csv};

    \addplot[yellow] table [x=speedup, y=f1, col sep=comma] {Dynamic-TinyBERTv2.csv};
    \addplot[purple] table [x=speedup, y=f1, col sep=comma] {Dynamic-TinyBERTv3.csv};
      \addplot[orange] table [x=speedup, y=f1, col sep=comma] {Dynamic-TinyBERTv4.csv};
      \addplot[gray, mark=*] table [x=speedup, y=f1, col sep=comma] {Dynamic-TinyBERT.csv};
    \addplot[black ,mark=*, mark size=3pt] coordinates{(1, 88.5)};
    \addplot[brown ,mark=*, mark size=3pt] coordinates{(1, 87.5)};
    \addplot[mark=none, black] coordinates {(1,88.5) (5.5,88.5)}; 
    \addplot[mark=none, brown] coordinates {(1,87.5) (5.5,87.5)}; 
    
    \legend{Dynamic-TinyBERT\textsubscript{w/ LD}naive, Dynamic-TinyBERT\textsubscript{w/ LD}v1, Dynamic-TinyBERT\textsubscript{w/ LD}v2, Dynamic-TinyBERT\textsubscript{w/ LD}v3, Dynamic-TinyBERT\textsubscript{w/ LD}v4, Dynamic-TinyBERT}
\end{axis}

\end{tikzpicture}
}

\hfill
\vspace{2mm}
\caption{Pareto-curves of Dynamic-TinyBERT models trained with LengthDrop vs. Dynamic-TinyBERT which was not trained with LengthDrop}
\label{fig:versions}

\end{figure}

\subsection{Dataset}
All our experiments are evaluated on the challenging question-answering benchmark SQuAD1.1~\cite{Rajpurkar2016SQuAD1Q}. Running a question-answering model with token elimination is possible only due to the Drop-and-Restore method which brings back the tokens in the last hidden layer. For simplicity, We did not use data augmentation at any stage, differently from the original TinyBERT which was trained partially on the original data and partially on augmented data. We leave experimenting with data-augmentation for future work.

\subsection{Setup}
We train all the described models on a Titan GPU. For our Dynamic-TinyBERT model we use the architecture of TinyBERT6L: a small BERT model with 6 layers, a hidden size of 768, a feed forward size of 3072 and 12 heads. To initialize the students we use the publicly released (second version) general TinyBERT\footnote{\url{https://github.com/huawei-noah/Pretrained-Language-Model/tree/master/TinyBERT}}. For all experiments we use sequence length of 384. For ID, whether with or without LengthDrop, we use lr=5e-5, batch size=16. For PD without LengthDrop we use lr=3e-5, batch size=16 and for PD with LengthDrop we use lr=2e-5, batch size=16. For training with LengthDrop, we set a LengthDrop probability and LayerDrop probability of 0.2 and set the number of sub-models (sandwiches) to be 2. For each model we find the Pareto frontier of the accuracy-efficiency tradeoff by running SigOpt HPO with range definition: $91 \leq x0 \leq 384$ and with the following constrains:
(1) $91 \leq x1 \leq x0\quad$(2)  $91 \leq x2 \leq x1$  \dots  (5) $91 \leq x5 \leq x4$. The low bound of 91 is according to the linear scale 0.2 drop ratio; The search range is set in the middle of each layer: $91 = (384 * 0.8^7 + 384 * 0.8^6) / 2$. The evaluation is done on a 2 socket 24 core CLX 6252N.

\subsection{Performance on CPU}
The Pareto curves of Dynamic-TinyBERT and Dynamic-TinyBERT\textsubscript{w/ LD}naive are shown in Figure~\ref{fig:dynamic}. The exact accuracy and speedup values of all tested model are presented in Table~\ref{tab:models-performance}. 
Dynamic-TinyBERT (as well as all other tested versions, see section~\ref{sec:length-drop}) achieves an accuracy-efficiency tradeoff superior to any other efficient approaches, running 2.7x faster than BERT-base with no accuracy loss and up to 3.3x faster than BERT-base with sequence-length reduction and minimal accuracy loss (<1\%). For comparison, the popular DistilBERT~\cite{Sanh2019DistilBERTAD} performs with 2.7\% loss-drop and only 1.7x speedup (both models holds 67M parameters). Dynamic-TinyBERT\textsubscript{w/ LD}naive implementation achieves a slightly better speedup than Dynamic-TinyBERT, running up to 3.54x faster than BERT-base with <1\% loss-drop, however it performs worse than Dynamic-TinyBERT for the higher computational budget points.

\subsection{Dynamic-TinyBERT with LengthDrop}
\label{sec:length-drop}
Figure~\ref{fig:versions} shows the Pareto-curves of Dynamic-TinyBERT-variations models trained with Length-Drop as well as the Dynamic-TinyBERT curve for comparison. Dynamic-TinyBERT\textsubscript{w/ LD}naive, Dynamic-TinyBERT\textsubscript{w/ LD}v1 and Dynamic-TinyBERT\textsubscript{w/ LD}v4 shows similar behaviour with a slightly better resilience to sequence-length reduction than Dynamic-TinyBERT, starting from the middle of the 1\%-loss area and lower, however all variations performs worse than Dynamic-TinyBERT – which was not trained with LengthDrop – within the top area of the 1\%-loss zone. This implies that, overall, LengthDrop did not add significant robustness to Dynamic-TinyBERT in this case but rather worsened its performance, except in relatively low computational budget use cases, in which LengthDrop provides non-significant added value in terms of speedup. 

The drop in accuracy of the models trained with LengthDrop is surprising, because LAT~\cite{kim-cho-2021-length} showed no performance drop but rather a significant rise in accuracy of the full model after training it with LengthDrop. We hypothesize that this behavior is due to either the large number of epochs required to train a good TinyBERT or to the difference in the fine-tuning method — LAT was originally tested on BERT and DistilBERT, which were fine-tuned using hard labels, while TinyBERT utilizes task specific distillation that uses soft labels for learning. We examined these hypotheses in a separate experiment by performing standard supervised fine-tuning of the general-TinyBERT model (see Figure~\ref{fig:training_fig}) rather than transformer-distillation for 5 epochs and then training it with LengthDrop for an additional 10 epochs , which indeed resulted in higher accuracy (albeit much lower than the accuracy gained by training with the transformer-distillation for many epochs). This experiment does not distinguish between our two hypotheses since it uses both a (relatively) low number of epochs and a method of learning from hard labels. Further research is required in order to clarify this matter.

\begin{table}[t]
\centering
\caption{Models performance analysis}
\label{tab:models-performance}
\begin{tabular}{@{}lllll@{}}

\toprule
 Model                 & Max F1 (full model) & Best Speedup within BERT-1\%        \\ \bottomrule
BERT-base   & 88.5          & 1x            \\ \midrule
DistilBERT & 85.8           & -               \\ \midrule
TinyBERT    & 87.5           &  2x                \\ \midrule
Dynamic-TinyBERT    & \textbf{88.71} &  3.3x          \\ \midrule
Dynamic-TinyBERT\textsubscript{w/ LD}naive  & 88.24          & 3.54x \\ \midrule
Dynamic-TinyBERT\textsubscript{w/ LD}v1  & 88.11 & \textbf{3.7x}          \\ \midrule
Dynamic-TinyBERT\textsubscript{w/ LD}v2  & 88.12     &  3.46x \\ \midrule
Dynamic-TinyBERT\textsubscript{w/ LD}v3  & 88.00          & 3.3x \\ \midrule
Dynamic-TinyBERT\textsubscript{w/ LD}v4  & 88.3        &  3.64x \\ \bottomrule
\end{tabular}


\end{table}

\section{Conclusions and future work}
In this paper, we propose Dynamic-TinyBERT, which leverages sequence-length reduction to further dynamically compress TinyBERT, allowing adaptive sequence-length sizes to accommodate different computational budget requirements with a best accuracy-efficiency tradeoff. Experiments on the SQuAD1.1 benchmark dataset demonstrate the effectiveness of our proposed Dynamic-TinyBERT compared with previous work on BERT compression, as well as the collapse of LengthDrop training to further improve Dynamic-TinyBERT's resilience to word-vectors elimination as shown by LAT~\cite{kim-cho-2021-length} to be successful on other models. In future work we intend to explore how to combine dynamic sequence-length with Sparsity and Low bit Quantization methods to achieve maximum throughput performance.

\bibliographystyle{abbrvnat}
\bibliography{references}

\end{document}